\documentclass[conference]{IEEEtran}
\IEEEoverridecommandlockouts
\usepackage{cite}
\usepackage{amsmath,amssymb,amsfonts}
\usepackage{algorithmic}
\usepackage{graphicx}
\usepackage{textcomp}
\usepackage{xcolor}
\def\BibTeX{{\rm B\kern-.05em{\sc i\kern-.025em b}\kern-.08em
    T\kern-.1667em\lower.7ex\hbox{E}\kern-.125emX}}
\begin{document}

\title{Can Large Language Models Self-Correct in Medical Question Answering? An Exploratory Study\\
\thanks{%
\textsuperscript{*} These authors contributed equally to this work.

This work was supported by the National Center for Complementary and Integrative Health [grant numbers R01AT009457, U01AT012871]; the National Institute on Aging [grant number R01AG078154]; the National Cancer Institute [grant number R01CA287413]; the National Institute of Diabetes and Digestive and Kidney Diseases [grant number R01DK115629]; and the National Institute on Minority Health and Health Disparities [grant number 1R21MD019134-01]. }
}

\author{
\IEEEauthorblockN{
Zaifu Zhan\textsuperscript{1,2,*},
Mengyuan Cui\textsuperscript{3,*},
Rui Zhang\textsuperscript{2},
}
\IEEEauthorblockA{\textsuperscript{1}\textit{Department of Electrical and Computer Engineering, University of Minnesota}, Minneapolis, USA}
\IEEEauthorblockA{\textsuperscript{2}\textit{Division of Computational Health Sciences, Department of Surgery, University of Minnesota}, Minneapolis, USA}
\IEEEauthorblockA{\textsuperscript{3} \textit{Department of Software Engineering, University of St. Thomas}, Saint Paul,USA}
\IEEEauthorblockA{zhan8023@umn.edu, cui06590@stthomas.edu, ruizhang@umn.edu}
}

\maketitle

\begin{abstract}
Large language models (LLMs) have achieved strong performance on medical question answering (medical QA), and chain-of-thought (CoT) prompting has further improved results by eliciting explicit intermediate reasoning; meanwhile, self-reflective (self-corrective) prompting has been widely claimed to enhance model reliability by prompting LLMs to critique and revise their own reasoning, yet its effectiveness in safety-critical medical settings remains unclear. In this work, we conduct an exploratory analysis of self-reflective reasoning for medical multiple-choice question answering: using GPT-4o and GPT-4o-mini, we compare standard CoT prompting with an iterative self-reflection loop and track how predictions evolve across reflection steps on three widely used medical QA benchmarks (MedQA, HeadQA, and PubMedQA). We analyze whether self-reflection leads to error correction, error persistence, or the introduction of new errors. Our results show that self-reflective prompting does not consistently improve accuracy and its impact is highly dataset- and model-dependent: it yields modest gains on MedQA but provides limited or negative benefits on HeadQA and PubMedQA, and increasing the number of reflection steps does not guarantee better performance. These findings highlight a gap between reasoning transparency and reasoning correctness, suggesting that self-reflective reasoning is better viewed as an analytical tool for understanding model behavior rather than a standalone solution for improving medical QA reliability.
\end{abstract}


\section{Introduction}
Large language models (LLMs)\cite{achiam2023gpt,devlin2019bert} have recently achieved substantial progress across a wide range of natural language processing tasks\cite{zhan2025evaluation,li2025benchmarking}, including information extraction\cite{zhan2025ramie}, question answering\cite{zhou2025automating}, reasoning\cite{hou2025benchmarking}, summarization\cite{fang2025multi}, and decision support\cite{vrdoljak2025review,zhan2026peer}. In the medical domain, these models have demonstrated strong performance on medical question answering (medical QA) benchmarks and related clinical NLP tasks\cite{zhou2025large}, prompting increasing interest in their potential use for clinical reasoning assistance, educational support, and downstream decision-making workflows. medical QA is inherently challenging because it requires a combination of broad biomedical knowledge, domain-specific terminology understanding, multi-step reasoning, and the ability to integrate evidence under uncertainty\cite{dong2025generative}. Moreover, because medicine is a safety-critical domain, the evaluation of LLMs must consider not only accuracy but also reliability, robustness, and trustworthiness, especially when models provide persuasive rationales that may influence human decisions.

Among commonly used benchmarks, MedQA-USMLE\cite{jin2021disease}, derived from the United States Medical Licensing Examination (USMLE), has emerged as a particularly demanding medical QA task. Derived from professional medical licensing examinations, MedQA-USMLE is designed to assess whether models can perform multi-step clinical reasoning, differential diagnosis, and hypothesis evaluation, rather than merely retrieving isolated facts. Compared with many general-domain multiple-choice QA datasets, medical QA often requires linking symptoms to pathophysiology, distinguishing between plausible competing diagnoses, and selecting the most appropriate management or next-step decision. As a result, MedQA-USMLE is frequently used as a representative benchmark for measuring an LLM's capability in clinical reasoning and medical decision-making, and strong performance on MedQA is often interpreted as evidence that a model can go beyond surface-level pattern matching. This also makes MedQA a valuable testbed for studying reasoning behaviors, failure modes, and error patterns in medical settings.

A key driver behind recent performance gains in complex reasoning tasks, including medical QA, has been the development of reasoning-oriented prompting strategies, most notably chain-of-thought (CoT) prompting\cite{Wei2022Cot}. By explicitly instructing LLMs to produce intermediate reasoning steps before committing to a final answer, CoT prompting has been shown to improve accuracy on tasks that require step-by-step reasoning. CoT has been widely studied in mathematical reasoning\cite{Wei2022Cot}, commonsense reasoning, and scientific question answering\cite{wang2024t}, and it has also been applied to medical QA benchmarks such as MedQA and similar datasets\cite{liu2024medcot,miao2024chain}. Beyond performance improvements, CoT prompting offers a practical benefit for interpretability and transparency\cite{zhang2025igniting}: it reveals a reasoning trace that can be inspected, compared, and analyzed, enabling researchers to diagnose where errors may occur. This characteristic has led to a growing research direction focused on reasoning analysis, where CoT is not only a technique for improving outputs but also a lens for understanding how LLMs arrive at decisions in high-stakes contexts.

As the field has advanced, researchers have increasingly recognized that single-pass reasoning is not always sufficient, motivating methods that extend CoT into iterative reasoning frameworks\cite{pang2024iterative,radha2024iteration,wu2025depth}. A prominent class of such approaches is self-reflective reasoning\cite{cao2025masr,lin2025v,huang2025medreflect}, also referred to as self-correction, self-critique, or iterative refinement. In these methods, an LLM is prompted to evaluate its own reasoning trace and answer, identify possible logical or factual issues, and revise its conclusion through one or more rounds of reflection. Self-reflection is attractive for several reasons. First, it operates purely at inference time, without requiring additional training data, fine-tuning, or parameter updates, making it broadly applicable to black-box models accessed via APIs. Second, it provides a simple mechanism to potentially improve reliability by allowing a model to reconsider its own output when uncertainty or inconsistency is detected. Third, it aligns with an intuitive hypothesis: if LLMs can generate coherent step-by-step explanations, they may also be capable of recognizing when those explanations are incomplete, contradictory, or medically implausible. Encouraging results reported in general-domain tasks have further strengthened the belief that self-reflection can improve final answer quality by enabling internal error correction.

However, despite increasing attention to self-reflective prompting, its effectiveness and reliability in medical question answering remain insufficiently characterized. Medical QA differs from many general-domain reasoning tasks in ways that may limit the benefits of self-reflection. Correct answers often depend on subtle clinical details, the precise interpretation of symptoms and findings, and the selection of the most appropriate option among several plausible candidates. In addition, medical questions frequently involve latent assumptions (e.g., patient demographics, acuity, contraindications) that must be inferred carefully. These properties raise important questions about whether self-reflection truly functions as an error-correction mechanism in safety-critical domains. If an LLM is capable of correcting its own mistakes through reflection, why does the initial chain-of-thought fail to produce the correct answer? Conversely, if the initial reasoning is flawed, what guarantees that subsequent self-evaluation will lead to genuine correction rather than reinforcing an incorrect hypothesis through post-hoc rationalization? In practice, repeated reflection could potentially amplify confident but incorrect reasoning, which is especially concerning in medical settings where persuasive explanations may increase user trust even when the answer is wrong. Therefore, it is essential to empirically validate when self-reflection helps, when it fails, and whether it introduces new failure modes.

Motivated by these concerns, we conduct an exploratory study of self-reflective reasoning in large language models for medical multiple-choice question answering. Rather than proposing a new method aimed at maximizing benchmark performance, our goal is to systematically analyze how LLM predictions evolve when models are explicitly prompted to review and revise their own chain-of-thought reasoning. We evaluate this behavior across three widely used medical QA benchmarks---MedQA, HeadQA, and PubMedQA---to capture a spectrum of clinical exam-style questions and biomedical literature-based reasoning scenarios. Specifically, we compare standard CoT prompting with an iterative self-reflection loop and track model answers across reflection steps. This allows us to study whether self-reflection leads to error correction, error persistence, or the introduction of new errors, and how these outcomes vary by dataset and model scale. By characterizing these behaviors, we aim to provide empirical insights into the feasibility, limitations, and reliability implications of self-reflective reasoning for medical QA in safety-critical applications.

\section{Methods}

\subsection{Method Overview}

In this work, we conduct an exploratory study on self-reflective reasoning in LLMs for medical multiple-choice question answering.
Our goal is not to propose a new performance-improving method, but to investigate whether LLMs are capable of identifying and correcting their own reasoning errors when explicitly prompted to reflect.

As shown in Figure~\ref{fig:method}, we first establish a CoT reasoning baseline, where the model generates an explicit step-by-step clinical rationale followed by an answer.
We then extend this baseline with an iterative self-reflection procedure that operates on the model’s own reasoning traces.
By comparing model behavior before and after reflection, we analyze whether self-reflection leads to error correction, error persistence, or the introduction of new errors.

\subsection{Datasets}

We evaluate our analysis on three widely used medical QA benchmarks that cover complementary sources of medical knowledge and reasoning demands. All datasets are evaluated using their official test splits, and we compute accuracy by comparing model predictions against the provided ground-truth answers. Dataset statistics are summarized in Tab.~\ref{tab:datasets}.
\begin{table}[htbp]
\centering
\caption{Datasets used in our study and their sources. We report only the test sets, as no training or fine-tuning is performed in this exploratory analysis.}
\begin{tabular}{lccc}
\hline
\textbf{Dataset} & \textbf{Test Set Size} & \textbf{Source} \\
\hline
MedQA(USMLE)        & 1,273 & US medical licensing examinations \\
HeadQA       & 244   & Spanish medical specialization exams \\
PubMedQA     & 500   & Biomedical literature (PubMed abstracts) \\
\hline
\end{tabular}
\label{tab:datasets}
\end{table}

\paragraph{MedQA}
MedQA(USMLE)\cite{jin2021disease} is a large-scale benchmark built from US medical licensing examinations, designed to assess professional-level medical knowledge and multi-step clinical reasoning. Questions typically require synthesizing clinical presentations, distinguishing between plausible differential diagnoses, and selecting the most appropriate answer among multiple choices. Due to its licensing-exam origin and emphasis on structured clinical reasoning, MedQA is commonly considered a challenging benchmark for evaluating medical reasoning capability in LLMs.

\paragraph{HeadQA}
HeadQA\cite{vilares2019headqa} consists of medical examination questions collected from Spanish medical specialization exams, covering diverse clinical specialties and realistic decision-making scenarios. Compared with MedQA, HeadQA provides a different distribution of topics and exam styles, offering an additional perspective on model generalization across medical QA settings. Importantly, HeadQA complements MedQA by testing whether observed self-reflection behaviors persist under a different exam source and clinical framing.

\paragraph{PubMedQA}
PubMedQA\cite{jin2019pubmedqa} is a biomedical question answering dataset constructed from PubMed abstracts, requiring reasoning grounded in scientific evidence rather than exam-style clinical heuristics. In contrast to MedQA and HeadQA, which are exam-based, PubMedQA emphasizes interpreting biomedical literature and linking questions to findings described in abstracts. This dataset therefore allows us to examine whether self-reflective prompting behaves differently when questions are more evidence-centric.

\subsection{Chain-of-Thought Baseline}
As a baseline, we employ CoT prompting to elicit explicit reasoning from the model.
Given a medical question, the model is instructed to generate a step-by-step clinical rationale and then select a final multiple-choice answer.
This baseline reflects standard reasoning-based prompting commonly used in prior work on medical question answering.

The CoT baseline serves two purposes in our study:
(1) it provides a transparent reasoning trace for each prediction, and
(2) it establishes a reference point against which we examine the effects of self-reflective reasoning.

\begin{figure*}[htbp]
    \centering
    \includegraphics[width=0.9\linewidth]{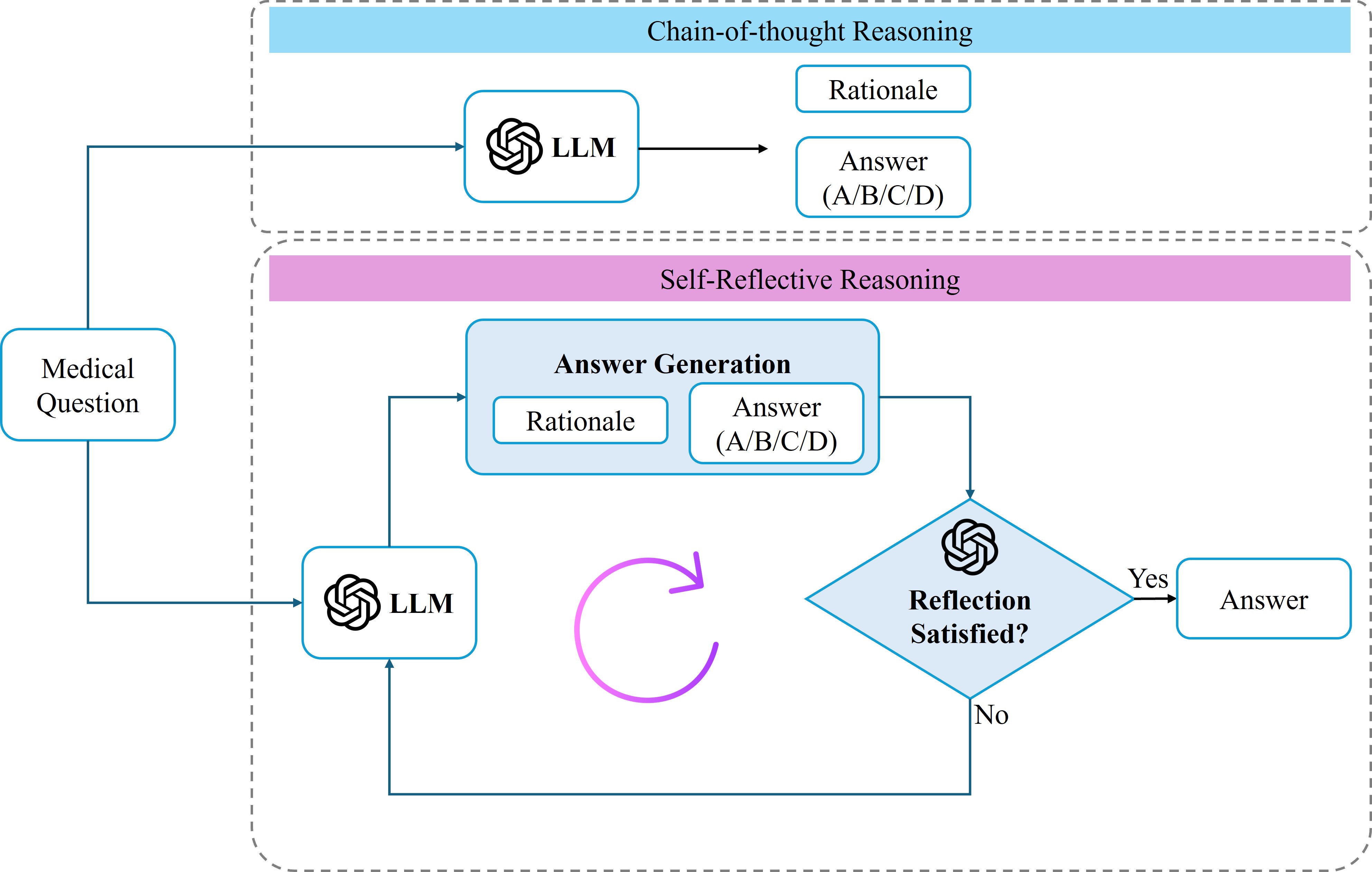}
    \caption{Overview of our self-reflective reasoning framework for medical multiple-choice question answering. Top: standard chain-of-thought (CoT) prompting, where the LLM produces an explicit rationale and a final option (A/B/C/D). Bottom: our self-reflective reasoning loop, which first generates an initial rationale--answer pair and then iteratively reviews the previous reasoning for medical or logical errors. If the reviewer is satisfied, the current answer is returned; otherwise, the model revises its rationale and updates the answer, repeating until convergence or a maximum number of reflection steps.}
    \label{fig:method}
\end{figure*}

\subsection{Self-Reflective Reasoning Framework}

We extend the CoT baseline with an iterative self-reflection framework, as shown in Fig. \ref{fig:method}, designed to probe the model’s ability to evaluate its own reasoning.

\paragraph{Initial Reasoning}
For each question, the model first produces a reasoning trace and an answer using the same CoT prompting as in the baseline.
This initial output is identical to the baseline prediction and ensures a fair comparison.

\paragraph{Iterative Self-Reflection}
After generating an initial answer, the model is prompted to act as a critical reviewer of its own reasoning.
In each reflection step, the model assesses whether the previous reasoning and answer are medically accurate and logically sound.
If the model determines that no errors are present, the reflection process terminates.
Otherwise, the model generates a revised reasoning trace and an updated answer.

The reflection loop is repeated for a fixed maximum number of steps.
Importantly, self-reflection is not assumed to improve accuracy; rather, it is used as an analytical tool to observe how the model responds when asked to evaluate and revise its own reasoning.

\paragraph{Answer Tracking and Aggregation}
We record the model’s predicted answer at each reflection step.
If the model fails to report satisfaction within the maximum number of reflection attempts, we additionally compute a majority-vote answer over all generated predictions.
This aggregation is used solely for analysis and does not assume that reflection improves performance.

\subsection{Models}
All experiments are conducted using large language models accessed via the OpenAI API.
Unless otherwise specified, we use GPT-4o\cite{achiam2023gpt} and GPT-4o-mini\cite{achiam2023gpt} as the primary model.
The models are treated as black-box systems without any parameter updates or fine-tuning.
All observed behaviors arise from prompt-based inference.

\subsection{Prompts}

We design two types of prompts corresponding to the CoT baseline and the self-reflection procedure. 

\paragraph{CoT Prompt}
The CoT prompt instructs the model to generate a step-by-step clinical rationale followed by a final answer.
The prompt encourages explicit reasoning but does not assume correctness.

\paragraph{Reflection Prompt}
The reflection prompt instructs the model to review its own previous reasoning and answer, and to determine whether errors are present.
If errors are identified, the model is asked to revise its reasoning and update its answer.
All outputs are constrained to a structured format to enable reliable analysis.

\begin{figure}[htbp]
    \centering
    \includegraphics[width=0.9\linewidth]{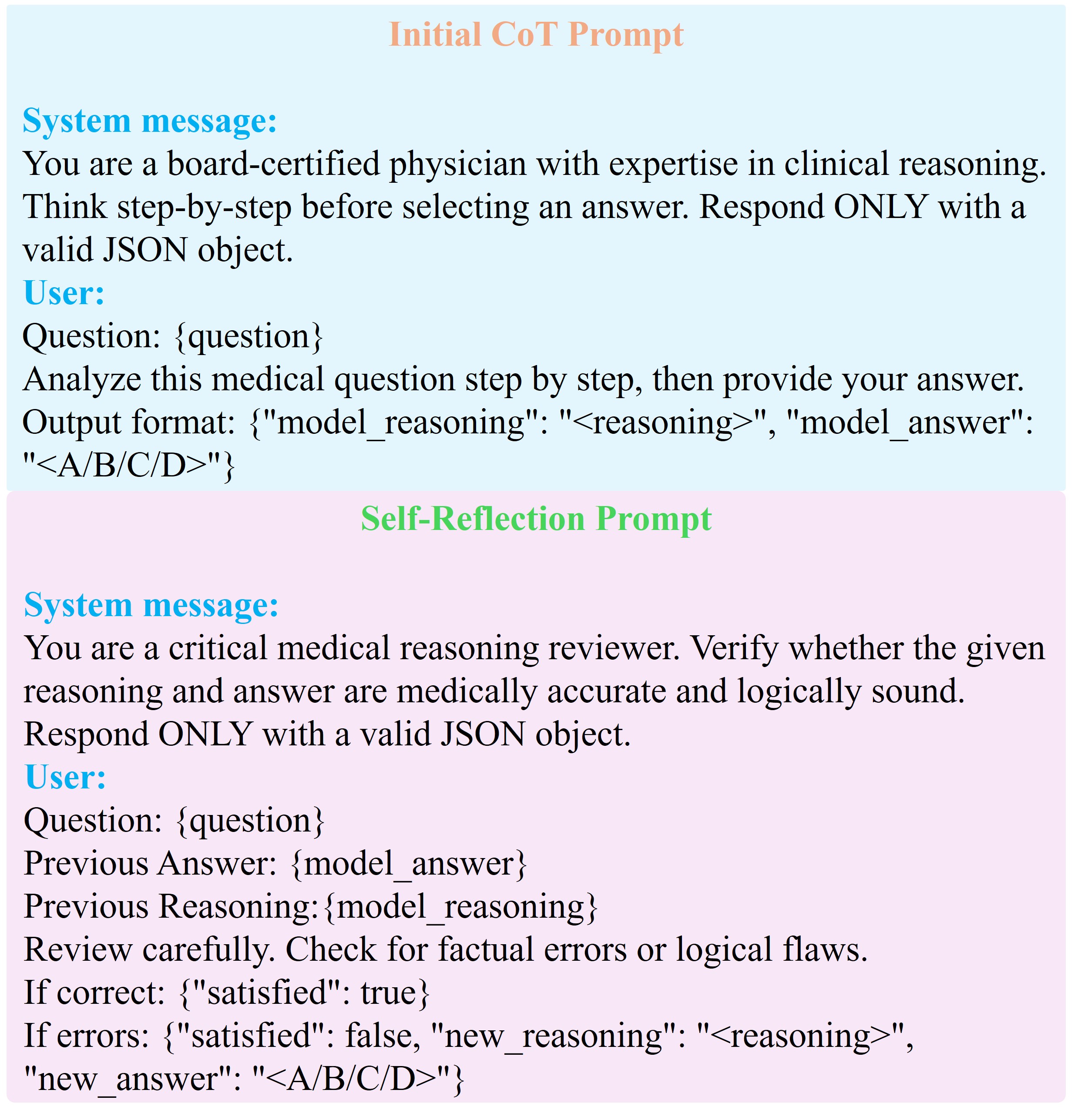}
    \caption{Example prompt used for the MedQA dataset. The prompt instructs the model to generate a step-by-step clinical rationale followed by a final multiple-choice answer, and is extended with a structured self-reflection instruction for iterative reasoning analysis.}
    \label{fig:promp}
\end{figure}

The specific prompt for the MedQA dataset is shown in Fig.~\ref{fig:promp}.

\subsection{Experimental Setup}

For each question, we allow up to $K$ self-reflection attempts, where $K=10$ unless otherwise specified.
We report accuracy for the CoT baseline as well as accuracy after each reflection step.

All experiments are conducted in a zero-shot setting without task-specific examples.
Rather than focusing on performance gains, we analyze how model predictions change across reflection steps, including cases where reflection corrects errors, fails to correct errors, or introduces new errors.

\begin{figure}[htbp]
    \centering
    \includegraphics[width=0.99\linewidth]{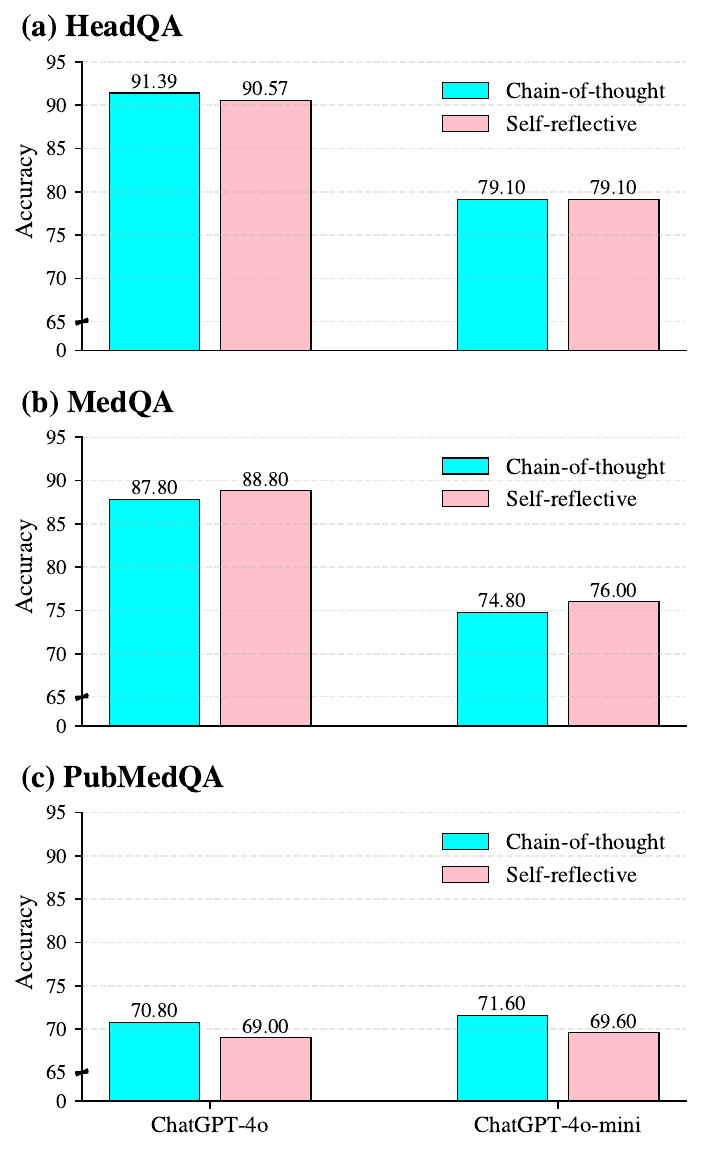}
    \caption{Comparison of Chain-of-Thought reasoning and self-reflective reasoning accuracy on three medical QA datasets (HeadQA, MedQA, and PubMedQA). Results are reported for ChatGPT-4o and ChatGPT-4o-mini.}
    \label{fig:cot_vs_reflect}
\end{figure}

\begin{figure*}[htbp]
    \centering
    \includegraphics[width=0.99\linewidth]{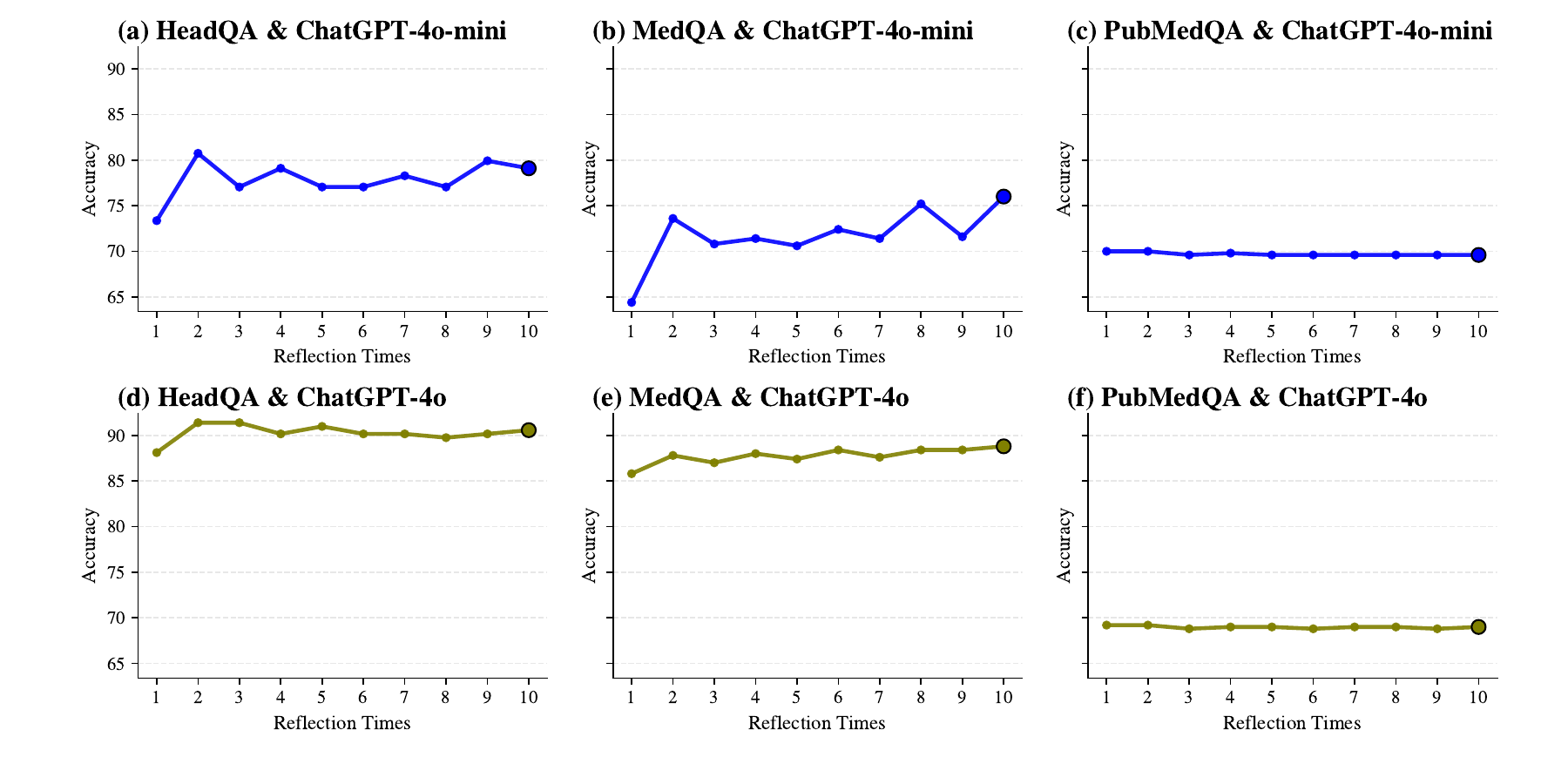}
    \caption{Accuracy as a function of the number of cumulative self-reflection steps for ChatGPT-4o and ChatGPT-4o-mini across three datasets. Each curve shows how performance evolves as additional reflection steps are applied.}
    \label{fig:acc_vs_steps}
\end{figure*}

\section{Results}

\subsection{Overall Comparison Between Chain-of-Thought and Self-Reflective Reasoning}
Fig.~\ref{fig:cot_vs_reflect} compares standard CoT prompting with our self-reflective reasoning procedure on three medical QA benchmarks (HeadQA, MedQA, PubMedQA) using ChatGPT-4o and ChatGPT-4o-mini. Overall, self-reflection does not consistently improve performance; instead, its effect depends on both the dataset and the model.

Specifically, on HeadQA, ChatGPT-4o slightly decreases after self-reflection (from 91.39\% to 90.57\%), while ChatGPT-4o-mini shows no change (79.10\% in both settings). On MedQA, self-reflection yields modest gains for both models, increasing accuracy from 87.80\% to 88.80\% for ChatGPT-4o and from 74.80\% to 76.00\% for ChatGPT-4o-mini. In contrast, on PubMedQA, self-reflection slightly degrades performance for both models (ChatGPT-4o: 70.80\% $\rightarrow$ 69.00\%; ChatGPT-4o-mini: 71.60\% $\rightarrow$ 69.60\%). These results indicate that self-reflection is not a uniformly beneficial inference strategy; in some cases, it may introduce additional errors.

\subsection{Accuracy Trends Across Reflection Steps}
To examine how performance evolves with increased reflection depth, Fig.~\ref{fig:acc_vs_steps} reports accuracy as a function of the cumulative number of reflection steps (up to 10 iterations). Across datasets, accuracy typically stabilizes after a small number of steps and does not exhibit monotonic improvement as reflection continues.

For HeadQA and MedQA, ChatGPT-4o remains relatively stable across reflection steps, with only minor fluctuations. ChatGPT-4o-mini shows larger variance, especially on MedQA, where accuracy oscillates across iterations rather than steadily improving. On PubMedQA, both models produce nearly flat curves, suggesting that additional reflection rarely changes the final answer. Overall, repeated self-reflection does not reliably accumulate benefits, and extended reflection may increase instability, particularly for the smaller model.

\subsection{Distribution of Reflection Depth}
\begin{figure*}[htbp]
    \centering
    \includegraphics[width=0.9\linewidth]{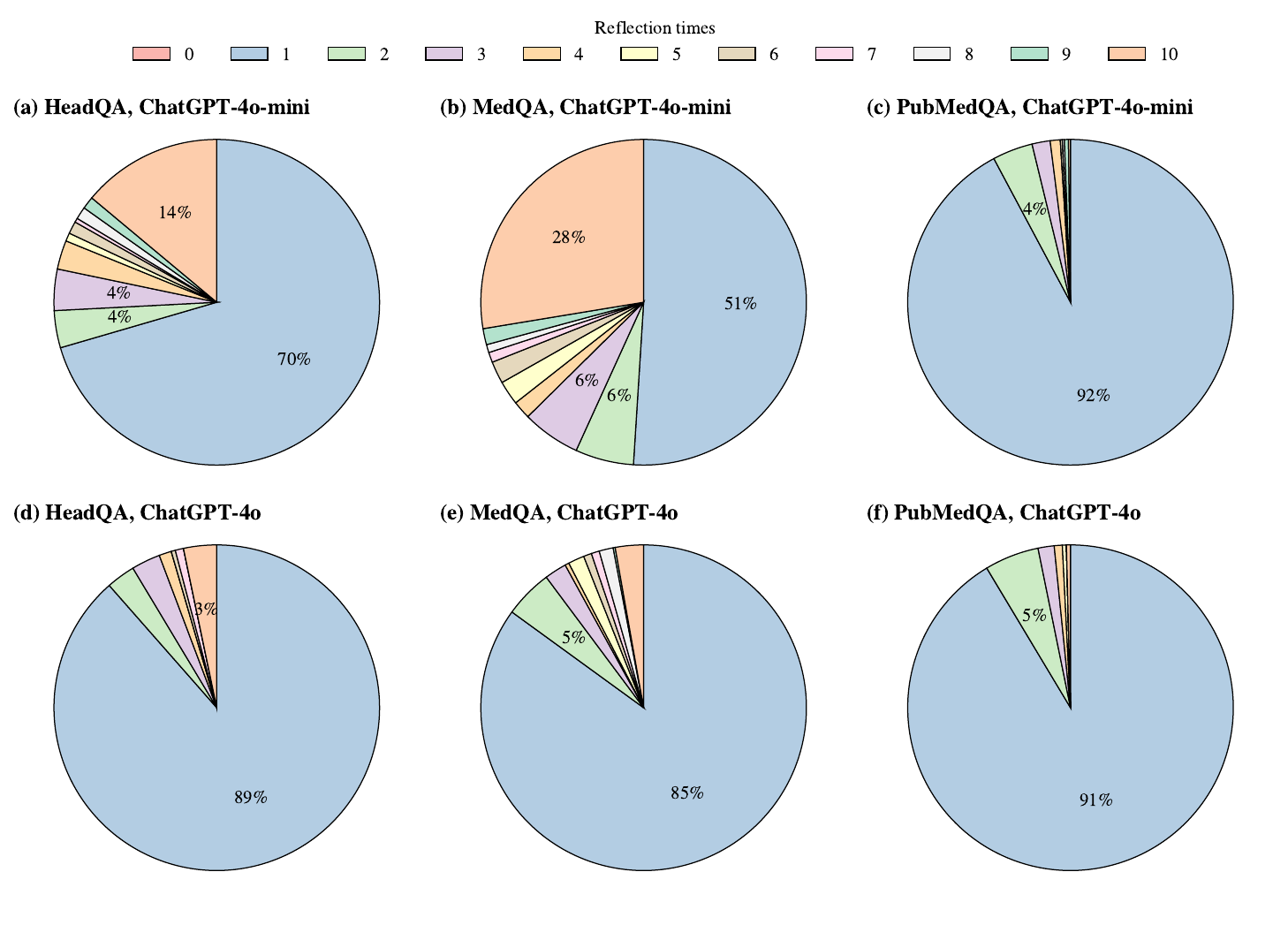}
    \caption{Distribution of the number of self-reflection steps used by the models across all evaluation instances. Each pie chart shows the percentage of samples requiring a given number of reflection steps (0–10) for a specific dataset–model pair. Colors are consistent across charts to facilitate comparison.}
    \label{fig:step_dist}
\end{figure*}

Fig.~\ref{fig:step_dist} shows the distribution of the number of reflection steps before the model reports satisfaction with its own reasoning. Across all datasets, most instances terminate with 0--1 reflection steps, implying that the model frequently judges its initial CoT reasoning to be sufficient.

For ChatGPT-4o, over 85\% of instances on each dataset require no additional reflection beyond the initial answer. ChatGPT-4o-mini engages in reflection more often, most notably on MedQA, where a substantial portion of examples require multiple reflection steps. However, deep reflection (near the maximum of 10 steps) remains uncommon in all settings. This pattern suggests that self-reflection is selectively triggered and that larger models are more decisive in their initial reasoning outputs.

\subsection{Average Reflection Depth Across Datasets}
\begin{figure}[htbp]
    \centering
    \includegraphics[width=0.9\linewidth]{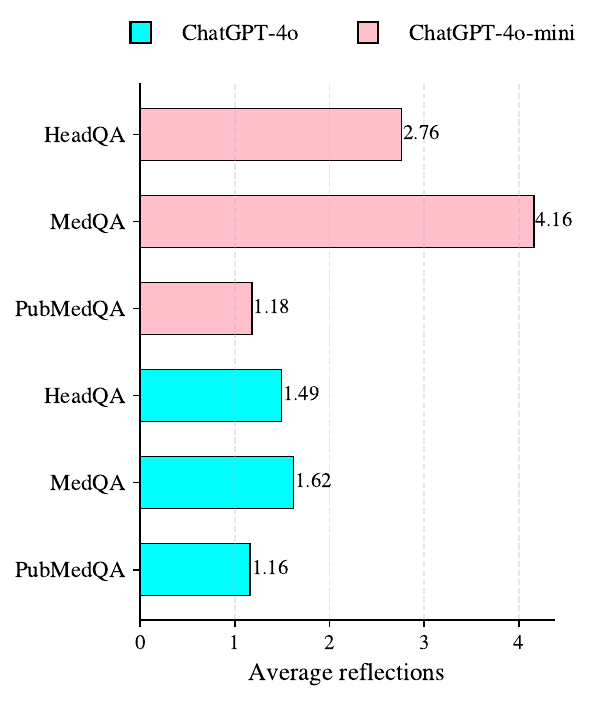}
    \caption{Average number of self-reflection steps used by ChatGPT-4o and ChatGPT-4o-mini on each dataset. Bars are colored by model, highlighting differences in reflection depth required across datasets.}
    \label{fig:avg_steps}
\end{figure}

Fig.~\ref{fig:avg_steps} summarizes the average number of reflection steps used by each model on each dataset. ChatGPT-4o consistently requires fewer reflection steps than ChatGPT-4o-mini, indicating higher stability in its initial reasoning. MedQA induces the highest average reflection depth for both models, whereas PubMedQA requires the fewest steps overall. The larger reflection depth on MedQA may reflect the complexity of professional-level exam questions that demand multi-step clinical reasoning, while PubMedQA appears less amenable to iterative self-correction through reflection.

\subsection{Summary of Findings}
Taken together, these results show that self-reflective reasoning is not a reliable performance enhancement mechanism for medical multiple-choice question answering. While reflection can occasionally correct initial errors, it can also preserve incorrect reasoning or introduce new errors. The impact of self-reflection is highly dependent on both dataset characteristics and model scale, and increasing the number of reflection steps does not guarantee improved accuracy.

\section{Discussion}
Medical question answering benchmarks such as MedQA play a central role in evaluating the clinical reasoning capabilities of large language models. Unlike fact-oriented biomedical QA datasets, MedQA is designed to reflect professional medical licensing examinations, requiring multi-step reasoning, differential diagnosis, and the integration of clinical knowledge under uncertainty. As a result, performance on MedQA is often used as a proxy for assessing whether a model’s reasoning behavior aligns with real-world clinical decision-making rather than surface-level pattern matching.

Given the growing adoption of self-reflective prompting strategies in medical and safety-critical applications, it is important to rigorously examine whether such techniques genuinely enhance model reliability. While self-reflection is frequently assumed to improve correctness by encouraging models to re-evaluate their own reasoning, this assumption remains underexplored in high-stakes medical settings. Validating when and why self-reflection succeeds or fails is therefore critical, not only for performance evaluation but also for understanding the limitations of deploying self-evaluating models in clinical contexts.

Our results provide empirical evidence that self-reflective reasoning, when applied as a purely prompt-based inference strategy, does not reliably improve medical multiple-choice question answering performance. Instead, its effects are highly dependent on dataset characteristics and model scale. These findings suggest that self-reflection should be viewed less as a general-purpose accuracy booster and more as a diagnostic tool for probing model reasoning behavior.

One possible explanation for the dataset-dependent behavior observed in our results lies in the nature of the reasoning required by different medical QA benchmarks. MedQA questions often involve multi-step clinical reasoning and differential diagnosis, where revisiting intermediate assumptions may occasionally help identify overlooked inconsistencies. In contrast, PubMedQA emphasizes evidence-based reasoning grounded in biomedical abstracts. In such cases, errors may stem from misinterpretation or incomplete utilization of evidence, which iterative self-reflection alone cannot easily correct. HeadQA, while also exam-based, tends to be more structured, leaving fewer opportunities for reflection-driven revision once an initial reasoning path has been formed.

Our findings also suggest that self-reflective prompting may not function as a true error-correction mechanism, but instead may reinforce existing reasoning trajectories. Once a model commits to an initial interpretation, subsequent reflection steps may serve to rationalize that decision rather than critically challenge it, resembling confirmation bias in human reasoning. This phenomenon is particularly concerning in medical settings, where incorrect conclusions accompanied by coherent and confident explanations may increase user trust despite being wrong. As a result, reasoning transparency alone should not be equated with reasoning reliability.

Additionally, we observe clear differences in self-reflective behavior across model scales. Larger models such as ChatGPT-4o tend to require fewer reflection steps and exhibit more stable performance, suggesting higher internal consistency in their initial chain-of-thought reasoning. Smaller models, while more prone to engaging in reflection, show greater variability across reflection steps, indicating that iterative self-evaluation may introduce instability rather than correction when underlying reasoning is weak. This highlights that the effectiveness of self-reflection is closely tied to model capacity and reasoning robustness.

Taken together, these observations underscore the limitations of treating self-reflective prompting as a standalone safety or reliability mechanism for medical AI systems. While self-reflection can provide useful insights into model behavior, it should be complemented with external grounding, verification, or uncertainty-aware mechanisms when applied to safety-critical domains such as medicine.
\subsection{Limitations and Future Work}

This study has several limitations that should be acknowledged. First, our analysis focuses exclusively on prompt-based self-reflective reasoning without incorporating external supervision, retrieval\cite{zhan2025mmrag}, or verification signals. As a result, the reflection process relies entirely on the model’s internal representations, which may reinforce incorrect reasoning rather than correct it. Second, we evaluate self-reflection primarily at the answer level, which may limit its ability to intervene in earlier reasoning steps where errors originate. Finally, our experiments are conducted on a limited set of multiple-choice medical QA benchmarks and do not extend to open-ended clinical reasoning or real-world decision support scenarios.

Future work may explore integrating self-reflective reasoning with external grounding mechanisms, such as retrieval-augmented evidence, uncertainty estimation, or expert-defined constraints, to guide the revision process more effectively. Additionally, applying reflection at intermediate reasoning stages, rather than only after a final answer is produced, may offer greater opportunities for meaningful correction. Extending this analysis to broader clinical tasks and more diverse evaluation settings could further clarify when and how self-reflection contributes to safer and more reliable medical AI systems.

\section{Conclusion}

This work presents an exploratory analysis of self-reflective reasoning in large language models for medical multiple-choice question answering. By comparing standard chain-of-thought prompting with an iterative self-reflection procedure across multiple medical QA benchmarks, we systematically examined whether prompting models to review and revise their own reasoning leads to more reliable predictions. Our findings show that self-reflective prompting does not consistently improve accuracy and that its impact is strongly dependent on dataset characteristics and model scale.

While modest improvements are observed on MedQA, self-reflection offers limited or negative benefits on HeadQA and PubMedQA, and increasing the number of reflection steps does not guarantee better performance. These results highlight a critical gap between reasoning transparency and reasoning correctness: although self-reflection encourages explicit reasoning and self-evaluation, it does not ensure error correction. Overall, our study suggests that self-reflective reasoning should be viewed as an analytical tool for understanding model behavior rather than a standalone solution for improving medical QA performance, motivating future work that combines reflection with external grounding or verification mechanisms.


\bibliographystyle{IEEEtran}
\bibliography{0_reference}

\end{document}